\documentclass[letterpaper]{article} 
\usepackage{aaai23}  
\usepackage{times}  
\usepackage{helvet}  
\usepackage{courier}  
\usepackage[hyphens]{url}  
\usepackage{graphicx} 
\usepackage{natbib}  
\usepackage{caption} 
\usepackage{algorithm}
\usepackage{algorithmic}

\usepackage{newfloat}
\usepackage{listings}
\DeclareCaptionStyle{ruled}{labelfont=normalfont,labelsep=colon,strut=off} 
\floatstyle{ruled}
\newfloat{listing}{tb}{lst}{}
\floatname{listing}{Listing}
\usepackage[group-separator={,}, group-minimum-digits=4]{siunitx}
\usepackage{booktabs}
\usepackage{multirow}
\usepackage{amsmath, amssymb}
\def\eg{\emph{e.g.}}

\def\ie{\emph{i.e.}}

\def\etc{\emph{etc}}

\def\etal{\emph{et al. }}  

\title{Semantics-Aware Dynamic Localization and Refinement for \\ Referring Image Segmentation}
\author{
    Zhao Yang,\textsuperscript{\rm 1}
    Jiaqi Wang,\textsuperscript{\rm 2}
    Yansong Tang,\textsuperscript{\rm 3}
    Kai Chen,\textsuperscript{\rm 2}
    Hengshuang Zhao,\textsuperscript{\rm 4}
    Philip H.S. Torr\textsuperscript{\rm 1}
}
\affiliations{
    \textsuperscript{\rm 1}University of Oxford,\\
    \textsuperscript{\rm 2}Shanghai AI Laboratory,\\
    \textsuperscript{\rm 3}Tsinghua-Berkeley Shenzhen Institute, Tsinghua University,\\
    \textsuperscript{\rm 4}The University of Hong Kong
}

\begin{document}

\maketitle

\begin{abstract}
Referring image segmentation segments an image from a language expression.
With the aim of producing high-quality masks, existing methods often adopt iterative learning approaches that rely on RNNs or stacked attention layers to refine vision-language features.
Despite their complexity, RNN-based methods are subject to specific encoder choices, while attention-based methods offer limited gains.
In this work, we introduce a simple yet effective alternative for progressively learning discriminative multi-modal features.
The core idea of our approach is to leverage a continuously updated query as the representation of the target object and at each iteration, strengthen multi-modal features strongly correlated to the query while weakening less related ones.
As the query is initialized by language features and successively updated by object features, our algorithm gradually shifts from being localization-centric to segmentation-centric.
This strategy enables the incremental recovery of missing object parts and/or removal of extraneous parts through iteration.
Compared to its counterparts, our method is more versatile---it can be plugged into prior arts straightforwardly and consistently bring improvements.
Experimental results on the challenging datasets of RefCOCO, RefCOCO+, and G-Ref demonstrate its advantage with respect to the state-of-the-art methods.
\end{abstract}

\section{Introduction}
\label{sec:introduction}

Given an image and a natural language expression that describes an object from the image, the task of referring image segmentation is to predict a pixel-wise mask that delineates that object~\cite{cheng2014imagespirit,hu2016segmentation}.
It has applications in a broad range of areas such as image editing, augmented reality, robotics, \etc.
Different from the more conventional semantic/instance segmentation tasks where targets fall into a pre-defined set of categories, this task requires an algorithm to not only predict an accurate mask for the target, but infer what the target is from a free-form language expression.

\begin{figure}[t]
\centering
\includegraphics[width=0.9\columnwidth]{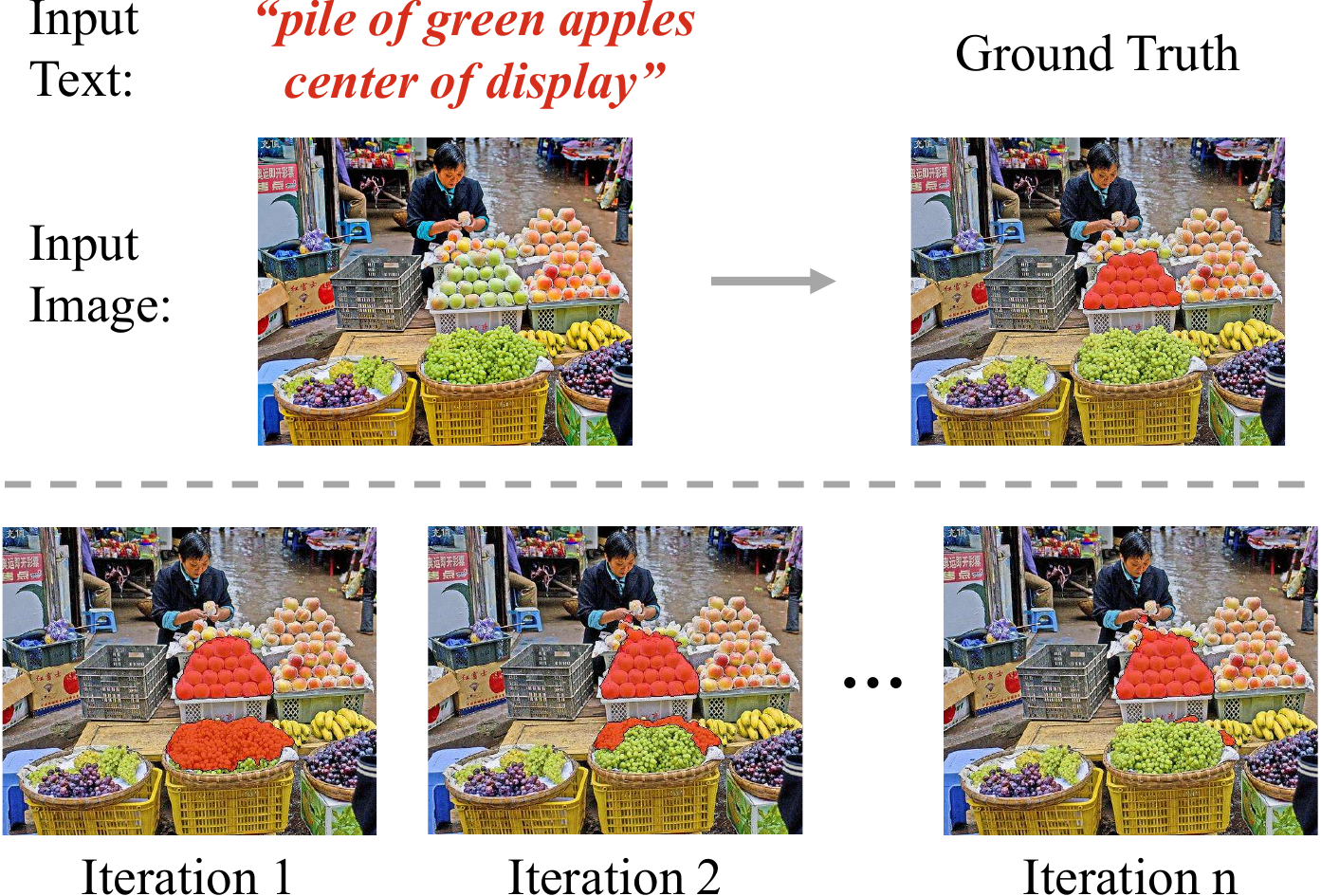}
\caption{Top half: Referring image segmentation aims to predict a mask delineating a target area described by a language expression. Bottom half: Our method tackles this problem in iterations, where it continuously improves upon its previous prediction based on accumulated object context, finally producing a highly accurate mask.}
\label{fig:1}
\end{figure}

Taking advantage of a fully convolutional architecture \cite{fcn}, current state-of-the-art methods~\cite{liu2017recurrentInter,li2018referring,chen2019see,luo2020cascade,liu2021cmpc,ding2021vlt,yang2021lavt} jointly address both requirements by developing powerful cross-modal feature fusion methods, which align linguistic meanings and visual cues in a common feature space.
Intuitively, learning accurate cross-modal alignments can be a daunting task as abundant noise is present in both the language and the vision inputs.
In this context, iterative learning becomes a characteristic approach that is adopted by many methods, which mitigates that difficulty: By performing cross-modal feature fusion in several rounds, alignments can be incrementally established and refined, subsequently leading to higher segmentation accuracy.

Existing state-of-the-art methods generally perform iterative learning via recurrent neural networks (RNNs) \cite{liu2017recurrentInter,li2018referring,chen2019see} or stacked attention layers \cite{luo2020cascade}.
RNN-based methods exploit the sequential property of the input data, either that of a sentence encoded step by step by a recurrent language model \cite{liu2017recurrentInter}, or that of an image encoded into a feature pyramid by a hierarchical vision backbone network \cite{li2018referring,chen2019see}.
Though having the potential of capturing helpful dependencies, they have several drawbacks.
Recurrent language models (\eg,~\cite{sutskever2014sequence}) are disadvantaged against the prevalent Transformer-based language models (\eg, \cite{bert}), while recurrently integrating multi-scale visual semantics only indirectly addresses cross-modal alignment.
Moreover, neither strategy is orthogonal to many of the other state-of-the-art fusion methods (such as ones proposed in VLT~\cite{ding2021vlt} and LAVT~\cite{yang2021lavt}).
The recent advance by Luo~\etal introduces cascade grouped attentions~\cite{luo2020cascade}, which refine cross-modal features of the whole image but lack the ability to exploit prior evidence for helping focus on regions that may need refinement the most---local regions where targets live and mistakes occur frequently.
As we observe in our experiments, this technique also does not bring improvements to baselines that adopt powerful Transformer-based fusion methods.

\begin{figure}[t]
\centering
\includegraphics[width=\columnwidth]{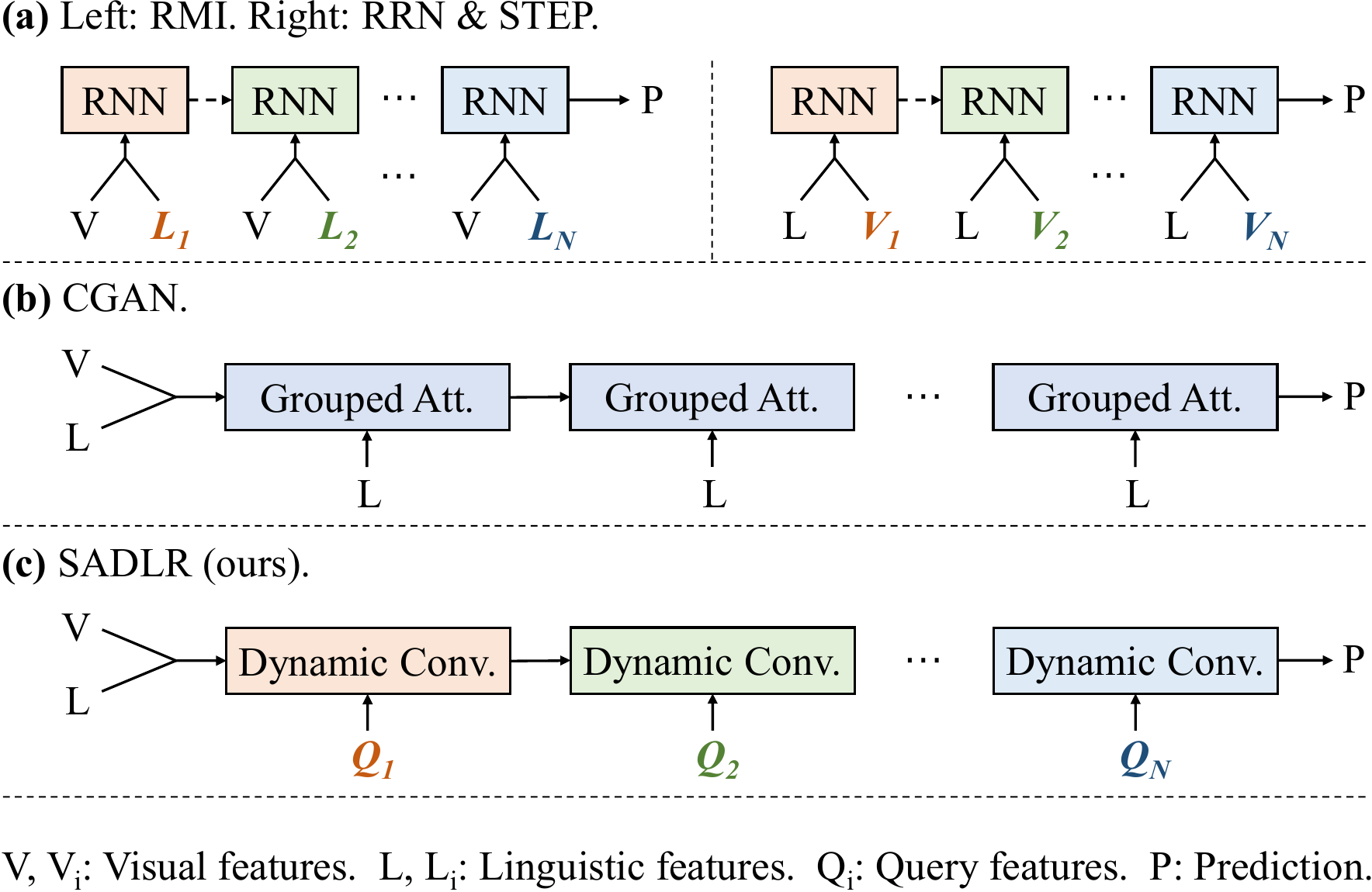}
\caption{A comparison of iterative learning schemes for referring image segmentation. (a)~The recurrent modeling approach exploits the sequential property of sentence features or the hierarchical property of pyramidal image features. (b)~The cascade attentions approach employs stacked attention layers where each layer enhances the vision-language mapping from the previous layer. (c)~Motivated differently, SADLR performs semantics-aware dynamic convolutions by leveraging continuously updated object context.}
\label{fig:2}
\end{figure}

To address those issues, we propose a semantics-aware dynamic localization and refinement (SADLR) method for referring image segmentation.
SADLR progressively strengthens target signals and suppresses background noise by leveraging a semantics-aware dynamic convolution module, which performs convolutions conditioning on a continuously updated representation of the target (what we denote as the ``query'').
To aid localization, in the first step, we initialize the query with a language feature vector summarized from the input expression, and predict a convolutional kernel from the query which operates on the multi-modal feature maps.
This step strengthens features according to language information.
In each of the following steps, we update the query with pooled object context obtained using the prediction from the previous step and perform dynamic convolution again.
As more object context is incorporated along the way, SADLR is able to gradually pinpoint the target and refine segmentation to a high degree of accuracy.

When applied to three state-of-the-art referring image segmentation methods, namely, LTS~\cite{jing2021locate}, VLT~\cite{ding2021vlt}, and LAVT~\cite{yang2021lavt}, the proposed SADLR approach is able to bring consistent performance improvements.
On the challenging benchmark datasets of RefCOCO~\cite{yu2016modeling}, RefCOCO+~\cite{yu2016modeling}, and G-Ref~\cite{nagaraja16refexp}, our method combined with the most powerful LAVT obtains 74.24\%, 64.28\%, and 63.60\% overall IoU on the validation sets, improving the state of the art for these datasets by absolute margins of 1.51\%, 2.01\%, and 2.36\%, respectively.

\section{Related Work}
\label{sec:related_work}

\noindent \textbf{Referring image segmentation} has attracted growing attention in the past decade.
While prior arts focus on developing cross-modal feature fusion methods based on CNN backbones \cite{hu2016segmentation,liu2017recurrentInter,chen2019see,feng2021efn,hu2020brinet,huang2020cmpc,hui2020linguistic,luo2020cascade,margffoy2018dynamic,shi2018key,ye2019cross}, recent developments leverage the Transformer architecture \cite{attention-all-you-need} for aggregating vision-language representations, including both at the feature encoding stage \cite{yang2021lavt} and at the feature decoding stage \cite{ding2021vlt}.

With the purpose of mitigating difficulties in cross-modal context modeling, some methods, including RMI~\cite{liu2017recurrentInter}, RRN~\cite{li2018referring}, STEP~\cite{chen2019see}, and CGAN~\cite{luo2020cascade}, adopt an iterative learning approach and refine cross-modal features in several rounds.
Mimicking how humans tackle visual grounding tasks, RMI integrates linguistic features with visual features step by step in a word-reading order, an approach well motivated for RNN language models but ill-suited to the fully-connected Transformer language models.
Conversely, RRN and STEP propose to integrate visual features with linguistic features stage by stage traversing through the image feature pyramid.
This approach mainly aims at capturing multi-scale semantics from the visual features and has been shown to be inferior to some of the other multi-stage fusion methods such as LSCM~\cite{hui2020linguistic} and CMPC~\cite{liu2021cmpc}.
The work most related to ours is CGAN, which leverages cascaded multi-head attention layers for iteratively refining cross-modal mappings of the whole image.
In contrast to our method, it does not seek to exploit predictions as priors to help focus refinement on more important regions.
Although in principle, the cascade attentions can be applied on top of existing models, we find in experiments that they only benefit the more conventional models which do not employ a powerful Transformer-based architecture for feature fusion.

\begin{figure*}[t]
\centering
\includegraphics[width=0.9\textwidth]{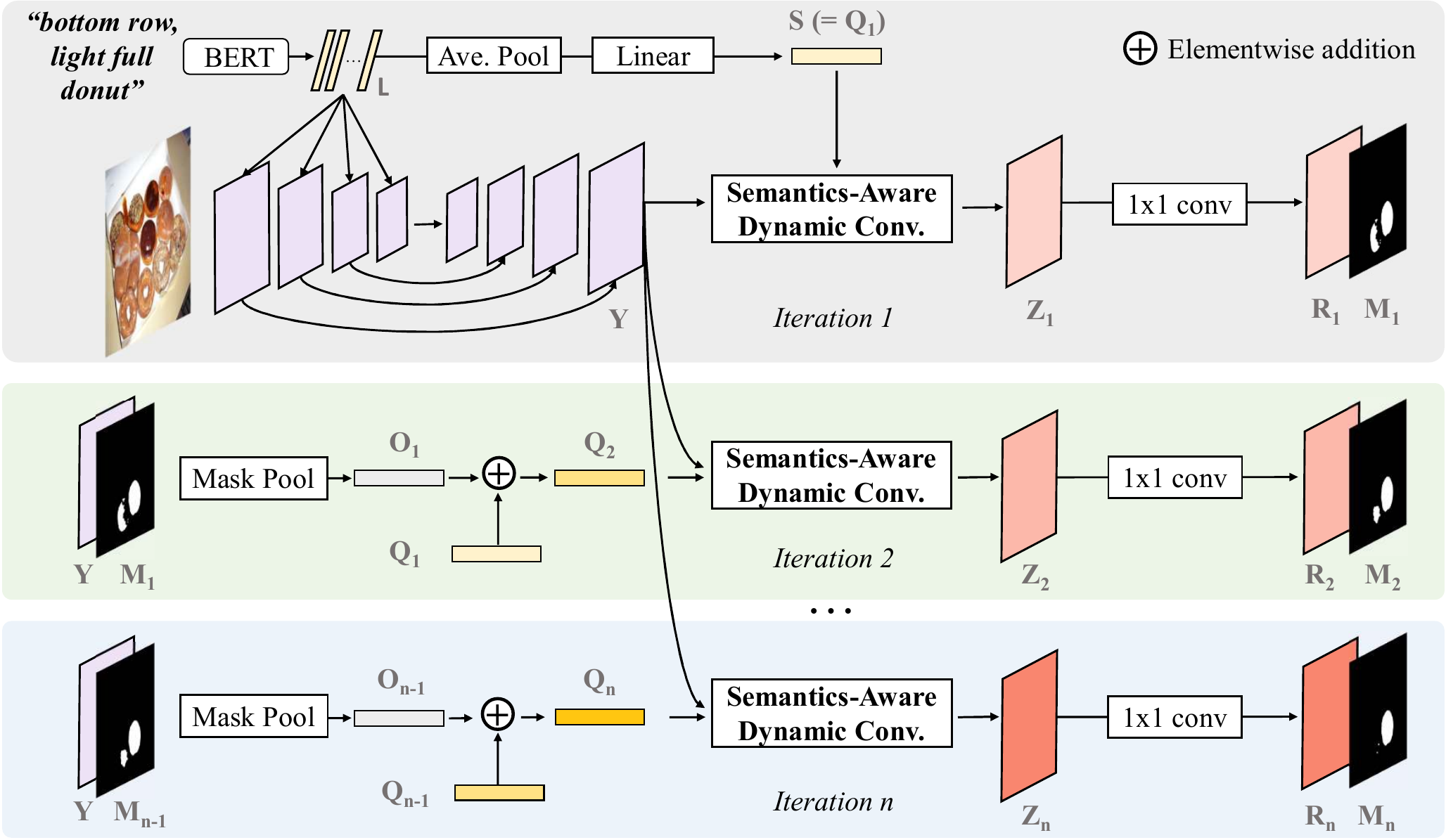}
\caption{A schematic illustration of the proposed SADLR approach.
In the first iteration, the query used for dynamic convolution is initialized by a language feature vector and segmentation mainly exploits language semantics.
In each of the following iterations, we continuously update our query with an object feature vector based on the previous iteration's prediction, and segmentation exploits an increasing amount of object semantics.
All parameters are shared across iterations.}
\label{fig:3}
\end{figure*}

Albeit our proposal is related to the above class of approaches, it offers important practical advantages.
First, it does not rely on a specific choice of the language encoder or the multi-modal feature encoding scheme.
Second, it is generalizable to a variety of top-performing models, including Transformer-based ones, as shown in our experiments.

\noindent \textbf{Dynamic convolution} is a versatile technique that can be applied for detecting user-defined patterns in a set of feature maps.
Originally proposed in the context of short-range weather prediction~\cite{klein2015dynamic}, it has recently found popularity in several query-based object detection systems~\cite{sun2021sparse,fang2021instances}.
In such models, a set of learnable embeddings serve as latent object features, and dynamic weights are predicted from these queries for localizing and classifying objects.
In comparison, we leverage dynamic convolution for solving the problem of segmentation from natural language.

Here, we make a distinction between the types of dynamic convolutions employed in this work and some earlier methods addressing the related task of referring video object segmentation~\cite{gavrilyuk2018actor,li2017tracking,wang2020context}.
In those methods, the term dynamic convolution/filtering refers to predicting classification weights that have the same number of channels as the target feature maps, from which a foreground score map is generated.
In contrast, our aim is to produce an enriched set of feature maps that highlight the semantics related to the dynamic query.

\noindent \textbf{Multi-modal reasoning} is a broader topic that covers many lines of work.
For instance, Radford \etal focus on the large-scale pre-training of transferable representations using paired image-text data~\cite{clip}.
Hu \etal propose the Unified Transformer model~\cite{hu2021unit} which jointly learns many vision-language tasks such as visual question answering~\cite{antol2015vqa} and visual entailment~\cite{xie2019visual}.
Kamath \etal devise MDETR~\cite{kamath2021mdetr}, an end-to-end modulated detector conditioning on language queries, which can be fine-tuned to perform downstream tasks such as few-shot learning for long-tailed detection~\cite{gupta2019lvis}.

\begin{figure*}[t]
\centering
\includegraphics[width=0.8\textwidth]{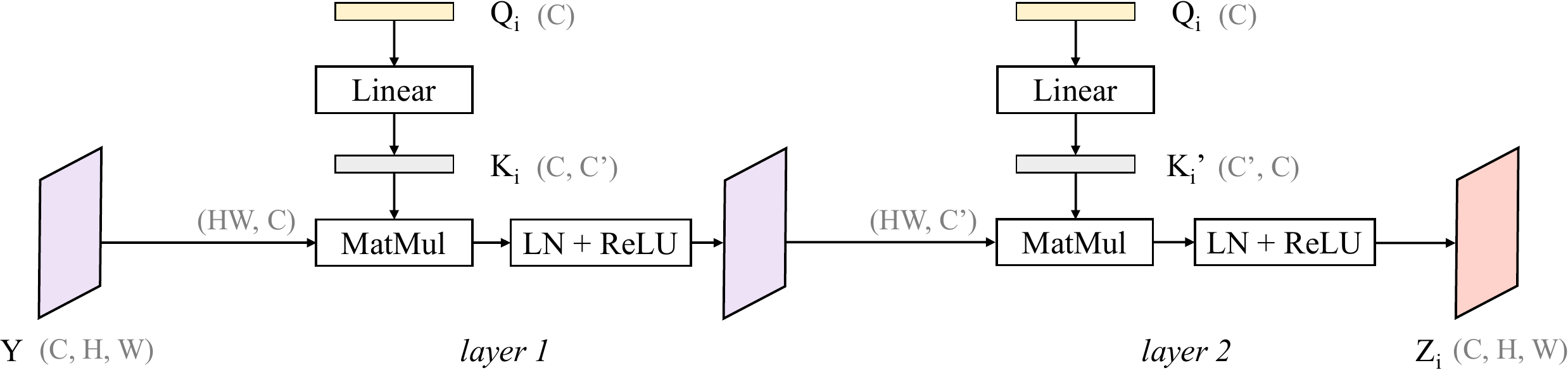}
\caption{Illustration of the semantics-aware dynamic convolution module at iteration $i$. The input feature maps $Y$ are strengthened by two layers of $1\times1$ convolution (implemented by matrix multiplication here), with output $Z_i$ capturing semantics encoded by the input query $Q_i$. `MatMul' denotes matrix multiplication and `LN' denotes layer normalization. This module is different from stacked (cross-)attention layers, for each of which, features at each output position would be computed as an aggregated sum of features at certain input positions based on pair-wise similarities (potentially with locality as a constraint).}
\label{fig:4}
\end{figure*}

\section{Method}
\label{sec:method}

\subsection{Overview}
\label{overview}
\noindent \textbf{Preliminaries}.
Fig.~\ref{fig:3} illustrates our method schematically.
The input consists of an image and a language expression.
We first extract a set of linguistic features, $L\in \mathbb{R}^{C_l\times N}$, from the input expression by leveraging a language encoder (such as BERT~\cite{bert} or an RNN~\cite{gru}).
Here, $C_l$ denotes the number of channels and $N$ denotes the number of words.
Then, the input image and the linguistic features $L$ go through a multi-modal feature encoding network, which can be instantiated by most any state-of-the-art referring segmentation model simply by removing its final classification layer.
Fig.~\ref{fig:3} illustrates our default choice of LAVT~\cite{yang2021lavt}, for which we include more details in Sec.~\ref{ssec:multi-modal_feature_encoding}.
This multi-modal feature encoding stage ensures that linguistic features and visual features are joined together in a single feature space that captures alignment information.
We denote the output set of multi-modal feature maps as $Y\in \mathbb{R}^{C\times H\times W}$, where $C$ is the number of channels and $H$ and $W$ denote the height and the width, respectively.

\noindent \textbf{Semantics-aware dynamic convolution}.
In the ensuing stage, instead of directly producing the object mask via a classifier, we iteratively refine the multi-modal features via a semantics-aware dynamic convolution module, which we further motivate and detail in Sec.~\ref{ssec:dynamic_convolution}.
Each layer of this module predicts a kernel from an input feature vector and convolves multi-modal feature maps with the predicted kernel.
In this paper, we call this kernel-generating feature vector a ``query,'' to give relevance to its functionality as an exemplar that characterizes the notion we wish to highlight in the multi-modal feature maps.

\noindent \textbf{Pipeline of each iteration}.
Assuming there are $n$ iterations in our refinement procedure, where $n$ is a hyper-parameter we study in Sec.~\ref{ssec:ablation_study},
we denote the dynamic query at iteration $i\in\{1,2,...,n\}$ as $Q_i\in \mathbb{R}^{C}$, where $C$ denotes the number of channels.
At each iteration, we send the query $Q_i$ and the multi-modal feature maps $Y$ (from the encoding stage) into the aforementioned semantics-aware dynamic convolution module, where a kernel is derived from $Q_i$ and convolved with $Y$ for producing a set of feature maps that highlight notions relevant to $Q_i$.
We denote the output feature maps as $Z_i\in \mathbb{R}^{C\times H\times W}$, where $H$ and $W$ denote the height and the width, respectively, and $C$ denotes the number of channels.
Next, a $1\times 1$ convolution projects $Z_i$ into raw score maps, denoted as $R_i\in \mathbb{R}^{2\times H\times W}$, where indices $0$ and $1$ along the channel dimension define the score maps for the ``background'' class and the ``object'' class, respectively.
We obtain a binary object mask $M_i\in \mathbb{R}^{H\times W}$ by taking \emph{argmax} along the channel dimension of $R_i$.

\noindent\textbf{Query initialization and iterative update}.
In different iterations, we use different queries with the purpose of highlighting different types of features.
To start with, at iteration $1$, we initialize $Q_1$ as a sentence feature vector, $S\in \mathbb{R}^{C}$, which is obtained from $L$ (from the language encoder) via average pooling across the word dimension followed by linear projection for channel reduction.
Since $Q_1$ contains pure linguistic information that summarizes the target object, the output from the semantics-aware dynamic convolution module $Z_1$ highlights referring localization information.
At iteration $i$, $i\in \{2,3,...n\}$, we update $Q_{i-1}$ with an object feature vector $O_{i-1}$ pooled from $Y$ using $M_{i-1}$ to obtain a new query $Q_{i}$ for the current iteration.
This process can be described mathematically as follows
\begin{align} \label{eq:1}
O_{i-1} &= \text{AvgPool}(M_{i-1}, Y), \\  \label{eq:2}
Q_i &= Q_{i-1} + O_{i-1},
\end{align}
where `AvePool' denotes computing a weighted average of feature vectors across the spatial dimensions of $Y$, with the binary mask $M_{i-1}$ being the weight map.
This query update scheme aims to progressively integrate more object context information into the query as the predicted mask $M_{i-1}$ grows more accurate by each iteration.
As mentioned earlier, dynamic convolution strengthens features related to the given query.
Therefore, as the query grows from being purely linguistic to containing more object-centric features, the algorithm gradually shifts focus from referring localization to segmentation refinement.
As shown in the qualitative analysis in Fig.~\ref{fig:5}, the predicted masks tend to grow more refined by each iteration.

\subsection{Multi-modal Feature Encoding} \label{ssec:multi-modal_feature_encoding}
By default, we adopt LAVT~\cite{yang2021lavt} as the multi-modal feature encoding network for its top performance in this task.
LAVT leverages a powerful hierarchical vision Transformer~\cite{liu2021swin} to jointly embed vision and language information.
At each stage of the vision backbone, a pixel-word attention module densely aligns linguistic features with visual features at each spatial location, and a language pathway directs integrated multi-modal cues into the next stage of the vision backbone.
Our proposed method is not limited to a specific multi-modal feature encoding network.
As we show in Table~\ref{tab:3}, when adopting LTS~\cite{jing2021locate} or VLT~\cite{ding2021vlt} as the encoding network, SADLR consistently obtains improved results with respect to the vanilla model.

\subsection{Semantics-Aware Dynamic Convolution} \label{ssec:dynamic_convolution}

\begin{table*}[t]
   \centering
   \setlength{\tabcolsep}{1.5mm}{
   \begin{tabular}{l|c|c|c|c|c|c|c|c|c}
      \toprule[1pt]
      \multirow{2}[4]{*}{}
      & \multicolumn{3}{c|}{RefCOCO} & \multicolumn{3}{c|}{RefCOCO+} & \multicolumn{3}{c}{G-Ref} \\
      & \multicolumn{1}{c}{val} & \multicolumn{1}{c}{test A} & \multicolumn{1}{c|}{test B} & \multicolumn{1}{c}{val} & \multicolumn{1}{c}{test A} & \multicolumn{1}{c|}{test B} & \multicolumn{1}{c}{val (U)} & \multicolumn{1}{c}{test (U)} & \multicolumn{1}{c}{val (G)}  \\
      \hline
      DMN~\cite{margffoy2018dynamic} & 49.78 & 54.83 & 45.13 & 38.88 & 44.22 & 32.29 & -     & -     & 36.76 \\
      RRN~\cite{li2018referring}     & 55.33 & 57.26 & 53.93 & 39.75 & 42.15 & 36.11 & -     & -     & 36.45 \\
      MAttNet~\cite{yu2018mattnet}   & 56.51 & 62.37 & 51.70 & 46.67 & 52.39 & 40.08 & 47.64 & 48.61 & -     \\
      CMSA~\cite{ye2019cross}        & 58.32 & 60.61 & 55.09 & 43.76 & 47.60 & 37.89 & -     & -     & 39.98 \\
      STEP~\cite{chen2019see}        & 60.04 & 63.46 & 57.97 & 48.19 & 52.33 & 40.41 & -     & -     & 46.40 \\
      BRINet~\cite{hu2020brinet}     & 60.98 & 62.99 & 59.21 & 48.17 & 52.32 & 42.11 & -     & -     & 48.04 \\
      CMPC~\cite{huang2020cmpc}      & 61.36 & 64.53 & 59.64 & 49.56 & 53.44 & 43.23 & -     & -     & 49.05 \\
      LSCM~\cite{hui2020linguistic}  & 61.47 & 64.99 & 59.55 & 49.34 & 53.12 & 43.50 & -     & -     & 48.05 \\
      CMPC+~\cite{liu2021cmpc}       & 62.47 & 65.08 & 60.82 & 50.25 & 54.04 & 43.47 & -     & -     & 49.89 \\
      MCN~\cite{luo2020mcn}        & 62.44 & 64.20 & 59.71 & 50.62 & 54.99 & 44.69 & 49.22 & 49.40 & -     \\
      EFN~\cite{feng2021efn}        & 62.76 & 65.69 & 59.67 & 51.50 & 55.24 & 43.01 & - & - & 51.93     \\
      BUSNet~\cite{yang2021busnet}     & 63.27 & 66.41 & 61.39 & 51.76 & 56.87 & 44.13 & - & - & 50.56 \\
      CGAN~\cite{luo2020cascade}     & 64.86 & 68.04 & 62.07 & 51.03 & 55.51 & 44.06 & 51.01 & 51.69 & 46.54 \\
      LTS~\cite{jing2021locate}       & 65.43 & 67.76 & 63.08 & 54.21 & 58.32 & 48.02 & 54.40 & 54.25 & - \\
      VLT~\cite{ding2021vlt}       & 65.65 & 68.29 & 62.73 & 55.50 & 59.20 & 49.36 & 52.99 & 56.65 & 49.76 \\
      CRIS~\cite{wang2021cris} & 70.47 & 73.18 & 66.10 & 62.27 & 68.08 & 53.68 & 59.87 & 60.36 & - \\
      LAVT~\cite{yang2021lavt} & 72.73 & 75.82 & 68.79 & 62.14 & 68.38 & 55.10 & 61.24 & 62.09 & 60.50 \\
      \hline
      Ours & \textbf{74.24} & \textbf{76.25} & \textbf{70.06} & \textbf{64.28} & \textbf{69.09} & \textbf{55.19} & \textbf{63.60} & \textbf{63.56} & \textbf{61.16} \\
      \bottomrule[1pt]
   \end{tabular}
   }%
   \caption{Comparison with other methods in terms of overall IoU. ``U'' and ``G'' represent UMD and Google, respectively.}
   \label{tab:1}%
\end{table*}%

Much like a traditional convolution, dynamic convolution convolves an input set of feature maps with the same kernel across all locations.
However, in dynamic convolution, the kernel is generated from a conditioning feature vector (what we call the query in this paper) instead of being fixed model parameters.
It particularly suits our need in referring image segmentation, as the generated kernel is sample-specific.

Fig.~\ref{fig:4} illustrates our semantics-aware dynamic convolution module.
It consists of two layers of dynamic convolutions.
At each iteration of SADLR, given query $Q_i$, a first linear function generates a dynamic kernel, $K_i\in \mathbb{R}^{C\times C'}$, where $C$ and $C'$ denote the input and the output numbers of channels, respectively.
Then convolution is performed via matrix multiplication between the input feature maps $Y$ and the dynamic kernel $K_i$, followed by layer normalization \cite{ba2016layer} and ReLU non-linearity \cite{nair2010rectified}.
Another convolutional kernel, $K_i'\in \mathbb{R}^{C'\times C}$, is subsequently generated from $Q_i$ by a second linear function.
Similarly, convolution is performed between the output from the previous layer and $K_i'$ via matrix multiplication, followed by layer normalization and ReLU non-linearity.

\subsection{Predicted Masks and the Loss Function}
\label{ssec:loss}
To produce segmentation masks, we upsample the raw score maps $R_i$ via bilinear interpolation to the resolution of the input image, and take the $argmax$ along the channel dimension of the upsampled score maps.
The loss function used for training is the following,
\begin{equation} \label{eq:3}
L = \lambda_1 L_1 + \lambda_2 L_2 + ... + \lambda_n L_n,
\end{equation}
where $n$ is the number of iterations, $L_i$, $i\in \{1,2,...n\}$, denotes the individual loss from iteration $i$, and $\lambda_i$, $i\in \{1,2,...n\}$, is the balancing weight for loss at iteration $i$.
Each individual loss is computed as the average Dice losses~\cite{milletari2016v} for the ``object'' class and the ``background'' class.
During inference, mask from the last iteration is used as the prediction.

\section{Experiments}
\label{sec:experiments}

\subsection{Datasets and Evaluation Metrics} \label{ssec:datasets_and_metrics}
\noindent \textbf{Datasets}.
We evaluate our proposed method on the datasets of RefCOCO~\cite{yu2016modeling}, RefCOCO+~\cite{yu2016modeling}, and G-Ref~\cite{mao2015generation,nagaraja16refexp}.
All three datasets collect images from the MS COCO dataset~\cite{lin2014mscoco}, and each object is annotated with multiple language descriptions from humans.
RefCOCO and RefCOCO+ are similar in size, both containing about 20K images, 50K annotated objects, and 140K expressions.
They are annotated in a two-player game, where each player is motivated to provide a minimally sufficient description for the other to identify the target.
As a result, expressions from these two datasets are succinct, averaging about 4 words per expression, and emphasize simple traits such as color, size, and position.
A special property about RefCOCO+ is that location words (such as ``left'' and ``right'') are banned, making it the harder of the two.
G-Ref contains roughly 27K images, 55K annotated objects, and 105K expressions.
Expressions in G-Ref tend to be longer, averaging around 8 words per expression, and provide more complete descriptions of the target objects.
As the expressions can be elaborate, G-Ref is the most challenging dataset of the three.
We evaluate our algorithm on both the UMD partition~\cite{nagaraja16refexp} and the Google partition~\cite{mao2015generation}.

\noindent \textbf{Evaluation metrics}.
We adopt three sets of metrics: precision@K (P@K, where K indicates an IoU threshold), mean IoU (mIoU), and overall IoU (oIoU).
The precision@K metric measures the percentage of test samples (\ie, image-sentence pairs) that pass an IoU threshold, where the IoU is computed between the prediction and the ground truth.
We evaluate precision at the thresholds of 0.5, 0.6, 0.7, 0.8, and 0.9.
On a separate note, this metric should be more accurately called ``recall@K.''
The mean IoU metric is the average IoU of all test samples.
This metric treats large and small objects equally.
The overall IoU is the intersection accumulated on all test samples divided by the union accumulated on all test samples.
It favors large objects.

\subsection{Implementation Details} \label{ssec:implementation_details}
As our best SADLR model is based on LAVT, most training and inference settings follow those in~\cite{yang2021lavt}.
We adopt the BERT-base model from~\cite{bert} as the language encoder, and the Swin-B model from~\cite{liu2021swin} as the backbone network.
BERT initialization weights are obtained from HuggingFace~\cite{wolf-2020-transformers}, and Swin initialization weights are ones pre-trained on ImageNet22K from~\cite{liu2021swin}.
The rest of the parameters in our model are randomly initialized.
The entire framework is trained end-to-end with the loss function defined in Eq.~\ref{eq:3}.
We adopt an AdamW~\cite{loshchilov2017adamw} optimizer with initial learning rate 5e-5 and weight decay 1e-2, and apply the ``poly'' learning rate scheduler~\cite{deeplabv3+}.
The numbers of channels $C_l$ and $C$ in Sec.~\ref{sec:method} are 768 and 512, respectively.
The default number of iterations ($n$ in Sec.~\ref{sec:method}) is 3, for which the loss weights, $\lambda_1$, $\lambda_2$, and $\lambda_3$, are 0.15, 0.15, 0.7, respectively.
For each $n$, we search for optimal balancing weights as follows.
While ensuring that the sum of the weights is 1, we experiment with large values ($\geqslant$0.5) on the last iteration, each time assigning equal weights to the rest of the iterations.
For each dataset, the model is trained on the training set for 40 epochs with batch size 32, where each object is sampled exactly once in an epoch (with one of its text annotations randomly sampled).
Images are resized to $480\times 480$ resolution and sentence lengths are capped at 20.

\subsection{Comparison with Others} \label{ssec:comparison_w_others}
In Table~\ref{tab:1}, we evaluate our proposed method against the state-of-the-art methods on the popular benchmarks of RefCOCO, RefCOCO+, and G-Ref.
Our approach obtains the highest overall IoU on all subsets of all benchmark datasets with various margins.
On the validation, test A, and test B subsets of RefCOCO, our method surpasses the second-best LAVT by margins of 1.51\%, 0.43\%, and 1.27\%, respectively.
On RefCOCO+, our method establishes a relatively large 2.01\% (absolute) lead with respect to the second-best method CRIS~\cite{wang2021cris} on the validation set, while also comparing favorably against the second-best LAVT on the test A and test B subsets with small advantages.
On both the validation and test sets of G-Ref (UMD partition), our approach improves the state of the art by solid margins of 2.36\% and 1.47\%, respectively.
On the most challenging validation set of G-Ref (Google partition), our method is also able to achieve new state-of-the-art results with a small improvement of 0.66\% (absolute).
In Table~\ref{tab:2}, we report the mean IoU of our method for comparison with Referring Transformer~\cite{muchen2021referring}, as mean IoU instead of overall IoU is reported in~\cite{muchen2021referring}.
Our method obtains better results on all three datasets.

\begin{table}[t]
   \centering
   \resizebox{\columnwidth}{!}{
   \setlength{\tabcolsep}{0.5mm}{
   \begin{tabular}{l|c|c|c|c|c|c|c|c|c}
      \toprule[1pt]
      \multirow{2}[4]{*}{} & \multicolumn{3}{c|}{RefCOCO} & \multicolumn{3}{c|}{RefCOCO+} & \multicolumn{3}{c}{G-Ref} \\
      & \multicolumn{1}{c}{val} & \multicolumn{1}{c}{test A} & \multicolumn{1}{c|}{test B} & \multicolumn{1}{c}{val} & \multicolumn{1}{c}{test A} & \multicolumn{1}{c|}{test B} & \multicolumn{1}{c}{val (U)} & \multicolumn{1}{c}{test (U)} & \multicolumn{1}{c}{val (G)}  \\
      \hline
      RefTrans & 74.34 & 76.77 & 70.87 & 66.75 & 70.58 & 59.40 & 66.63 & 67.39 & - \\
      Ours & \textbf{76.52} & \textbf{77.98} & \textbf{73.49} & \textbf{68.94} & \textbf{72.71} & \textbf{61.10} & \textbf{67.47} & \textbf{67.73} & \textbf{65.21} \\
      \bottomrule[1pt]
   \end{tabular}
   }
   }
   \caption{Comparison with Referring Transformer~\cite{muchen2021referring} by mean IoU.}
   \label{tab:2}%
\end{table}%

\subsection{Ablation Study} \label{ssec:ablation_study}
In this section, we evaluate the effectiveness of the SADLR method and study the effects of several key design choices.
The experiment setups, including the language encoder, the visual backbone network, the loss, \etc., are kept the same in all studies for fair comparisons.
As a result, there is a difference between our reproduced overall IoU of LAVT (reported in Tables~\ref{tab:3} and~\ref{tab:4}) and that of LAVT from the original paper (reported in Table~\ref{tab:1}).
This difference mainly originates from the Dice loss that we adopt in all the experiments---the original LAVT adopts a cross-entropy loss.

\begin{table}[t]
   \centering
   \resizebox{\columnwidth}{!}{
   \setlength{\tabcolsep}{0.5mm}{
   \begin{tabular}{l|ccccc|cc|cc}
      \toprule[1pt]
      Method & P@0.5 & P@0.6 & P@0.7 & P@0.8 & P@0.9 & oIoU & mIoU & FLOPs & Time\\
      \specialrule{0.7pt}{1pt}{1pt}
      LTS & 82.10 & 77.51 & 70.94 & 58.54 & 27.14 & 70.84 & 71.92 & 133.3G & 41.0 \\
      LTS + CGAN & 84.52 & 80.72 & 74.89 & 61.57 & 29.13 & 71.39 & 73.57 & 143.3G & 48.8 \\
      LTS + SADLR & \textbf{85.20} & \textbf{81.79} & \textbf{76.39} & \textbf{64.92} & \textbf{32.28} & \textbf{72.22} & \textbf{74.49} & 139.4G & 46.8 \\
      \specialrule{0.7pt}{1pt}{1pt}
      VLT & 83.80 & 79.96 & 74.10 & 60.52 & 25.60 & 70.31 & 72.62 & 142.6G & 41.9 \\
      VLT + CGAN & 80.33 & 76.33 & 69.94 & 56.32 & 22.80 & 66.41 & 69.62 & 152.7G & 49.0 \\
      VLT + SADLR & \textbf{85.00} & \textbf{81.85} & \textbf{76.17} & \textbf{62.44} & \textbf{27.81} & \textbf{71.64} & \textbf{73.90} & 148.6G & 47.7 \\
      \specialrule{0.7pt}{1pt}{1pt}
      LAVT & 85.49 & 82.00 & 76.44 & 65.65 & 35.16 & 73.12 & 75.30 & 197.4G & 45.6 \\
      LAVT + CGAN & 85.14 & 81.37 & 75.98 & 64.57 & 34.35 & 72.56 & 75.05 & 207.5G & 54.9 \\
      LAVT + SADLR & \textbf{86.90} & \textbf{83.68} & \textbf{78.76} & \textbf{67.93} & \textbf{37.36} & \textbf{74.24} & \textbf{76.52} & 203.5G & 51.5 \\
      \bottomrule[1pt]
   \end{tabular}
   }
   }
   \caption{Comparison between our method, CGAN, and several baseline methods on the RefCOCO validation set. The ``\emph{baseline} + SADLR'' or ``\emph{baseline} + CGAN'' notation denotes that we apply SADLR or CGAN on top of the baseline model. As described in the text below, all experiments adopt BERT as the language encoder and Swin-B as the backbone network, and follow the same training settings.}
   \label{tab:3}%
\end{table}%

\noindent \textbf{Comparison with CGAN and baseline methods}.
In Table~\ref{tab:3}, we apply our iterative dynamic refinement procedure on three different state-of-the-art methods, namely, LTS~\cite{jing2021locate}, VLT~\cite{ding2021vlt}, and LAVT~\cite{yang2021lavt}, and show that our method can generalize to all three architectures.
When using either the Transformer-based VLT or LAVT as the baseline method (the multi-modal feature encoding network in Sec.~\ref{sec:method}), our method consistently brings 1 to 2 absolute points of improvement on all of the seven evaluation metrics, including precision at the five threshold values, the overall IoU and the mean IoU.
On the non-Transformer-based LTS, our proposal achieves still greater improvements.
From P@0.5 to P@0.9, our method achieves 3.10\%, 4.28\%, 5.45\%, 6.38\%, and 5.14\% absolute improvements, respectively.
The gain generally increases as the threshold gets larger, which highlights the efficacy of our method in refining segmentation masks to a high degree of accuracy.
Moreover, on the overall IoU and mean IoU, our method attains 1.38\% and 2.57\% absolute improvements, respectively.
Overall, the results above highlight the generality and effectiveness of SADLR.

On the contrary, CGAN~\cite{luo2020cascade} is only able to bring moderate improvement to the more conventional LTS model, which is the best model among ones not employing Transformers for cross-modal feature fusion.
When the more effective Transformer architecture has been employed, applying CGAN leads to performance deterioration, as shown by the ``VLT + CGAN'' and ``LAVT + CGAN'' variants.

To analyze the computational overhead of SADLR, we report the FLOPs and the inference time (in milliseconds) of all models in Table~\ref{tab:3}.
We measure the inference time by averaging over 500 forward passes using batch size 1 at $480\times 480$ input resolution on an NVIDIA Quadro RTX 8000.
SADLR brings 3.09\%, 4.58\%, and 4.21\% relative increase of FLOPs and 12.94\%, 14.15\%, and 13.84\% relative increase of inference time to LAVT, LTS, and VLT, respectively, all of which are smaller than those of CGAN.

\noindent \textbf{Number of iterations}.
Next, we study the optimal number of iterations in SADLR.
As shown in Table~\ref{tab:4} (a), as the number of iterations grows, results continuously improve, with optimal ones obtained when there are three iterations.
Adding a fourth iteration generally does not yield better results (except marginally so on P@0.8).
Applying 0 iteration equals to using the baseline method (and in these experiments, LAVT).
These observations are consistent with our intuition that refinement may eventually saturate but before saturation, applying more iterations is beneficial.

\begin{table}[t]
   \centering
   \resizebox{\columnwidth}{!}{
   \setlength{\tabcolsep}{0.75mm}{
   \begin{tabular}{l|c|c|c|c|c|c|c}
      \toprule[1pt]
      & P@0.5 & P@0.6 & P@0.7 & P@0.8 & P@0.9 & oIoU & mIoU\\
      \bottomrule[1pt]
      \multicolumn{8}{l}{ \textbf{(a)} number of iterations} \\
      \toprule[1pt]
      0 & 85.49 & 82.00 & 76.44 & 65.65 & 35.16 & 73.12 & 75.30 \\
      1 & 85.44 & 82.01 & 76.68 & 65.92 & 35.86 & 73.22 & 75.41 \\
      2 & 86.63 & 83.63 & 78.48 & 67.83 & 37.04 & 73.99 & 76.32 \\
      3 (*) & \textbf{86.90} & \textbf{83.68} & \textbf{78.76} & 67.93 & \textbf{37.36} & \textbf{74.24} & \textbf{76.52} \\
      4 & 86.64 & 83.12 & 77.77 & \textbf{68.01} & 36.87 & 73.56 & 76.09 \\
      \bottomrule[1pt]
      \multicolumn{8}{l}{ \textbf{(b)} structure of the semantics-aware dynamic convolution module} \\
      \toprule[1pt]
      $[128]$ & 86.26 & 82.84 & 77.74 & 67.36 & 36.69 & 72.98 & 75.87 \\
      $[256]$ & 86.50 & 83.27 & 78.38 & 67.37 & 37.16 & 73.62 & 76.22 \\
      $[64,256]$ & \textbf{87.00} & \textbf{83.88} & 78.33 & 67.75 & 37.02 & 73.76 & 76.34 \\
      $[512]$ & 86.33 & 83.21 & 78.27 & 67.53 & 36.73 & 73.56 & 76.02 \\
      $[128,512]$ (*)& 86.90 & 83.68 & \textbf{78.76} & \textbf{67.93} & \textbf{37.36} & \textbf{74.24} & \textbf{76.52} \\
      $[128, 512] \times 2 $ & 86.47 & 83.27 & 78.09 & 67.19 & 36.87 & 73.70 & 76.12 \\
      \bottomrule[1pt]
      \multicolumn{8}{l}{ \textbf{(c)} query update method} \\
      \toprule[1pt]
      sum (*) & \textbf{86.90} & \textbf{83.68} & \textbf{78.76} & \textbf{67.93} & \textbf{37.36} & \textbf{74.24} & \textbf{76.52} \\
      replace & 86.52 & 83.48 & 78.36 & 67.09 & 37.35 & 73.19 & 76.14 \\
      \bottomrule[1pt]
   \end{tabular}
   }
   }
   \caption{Ablation studies on the RefCOCO validation set. Rows with (*) indicate our default choices. In each experiment, we change only one variable from our default configuration and study its effect on the overall model.}
   \label{tab:4}%
\end{table}%

\noindent \textbf{Structure of the semantics-aware dynamic convolution module}.
Our default implementation of this module consists of two convolution layers.
In Table~\ref{tab:4} (b), we study the effects of the number of convolution layers and the number of filters (the number of output channels) for each layer.
We define $[a]$ as one convolution layer with $a$ filters and $[a,b]$ as two convolution layers with $a$ and $b$ filters, respectively, where $a$ and $b$ are numbers.
Similarly, we define $[a,b]\times k$ as a stack of $k$ $[a,b]$ sub-structures, where $k$ is a number.
As shown in Table~\ref{tab:4} (b), the structure $[128, 512]$ produces the best results in five of the seven metrics, and is chosen as our default implementation.
$[64, 256]$ comes as the overall second-best structure and produces the best P@0.5 and P@0.6.
We note that when using only one layer of convolution, having 256 filters works better than having 512 filters (shown via a comparison between the ``$[256]$'' and ``$[512]$'' variants).
Conversely, when using two layers of convolutions, having more filters prevails over having fewer filters (shown via a comparison between the ``$[128,512]$'' and ``$[64,256]$'' variants).
This suggests that adopting more filters can be beneficial, but only likely so when more layers are employed which facilitates learning.
In addition, we show that repeating our best $[128,512]$ structure once (leading to a total of four convolution layers) does not bring further improvements.

\noindent \textbf{Query update method}.
In Table~\ref{tab:4} (c), we compare summation with an alternative replacement strategy when updating the query across iterations.
This corresponds to changing Eq.~\ref{eq:2} to $Q_i=O_{i-1}$, where the object feature vector replaces the previous query as the new one.
Table~\ref{tab:4} (c) shows that replacement leads to inferior results, which indicates that it is beneficial to retain language information in the query.

\begin{figure}[t]
\centering
\includegraphics[width=\columnwidth]{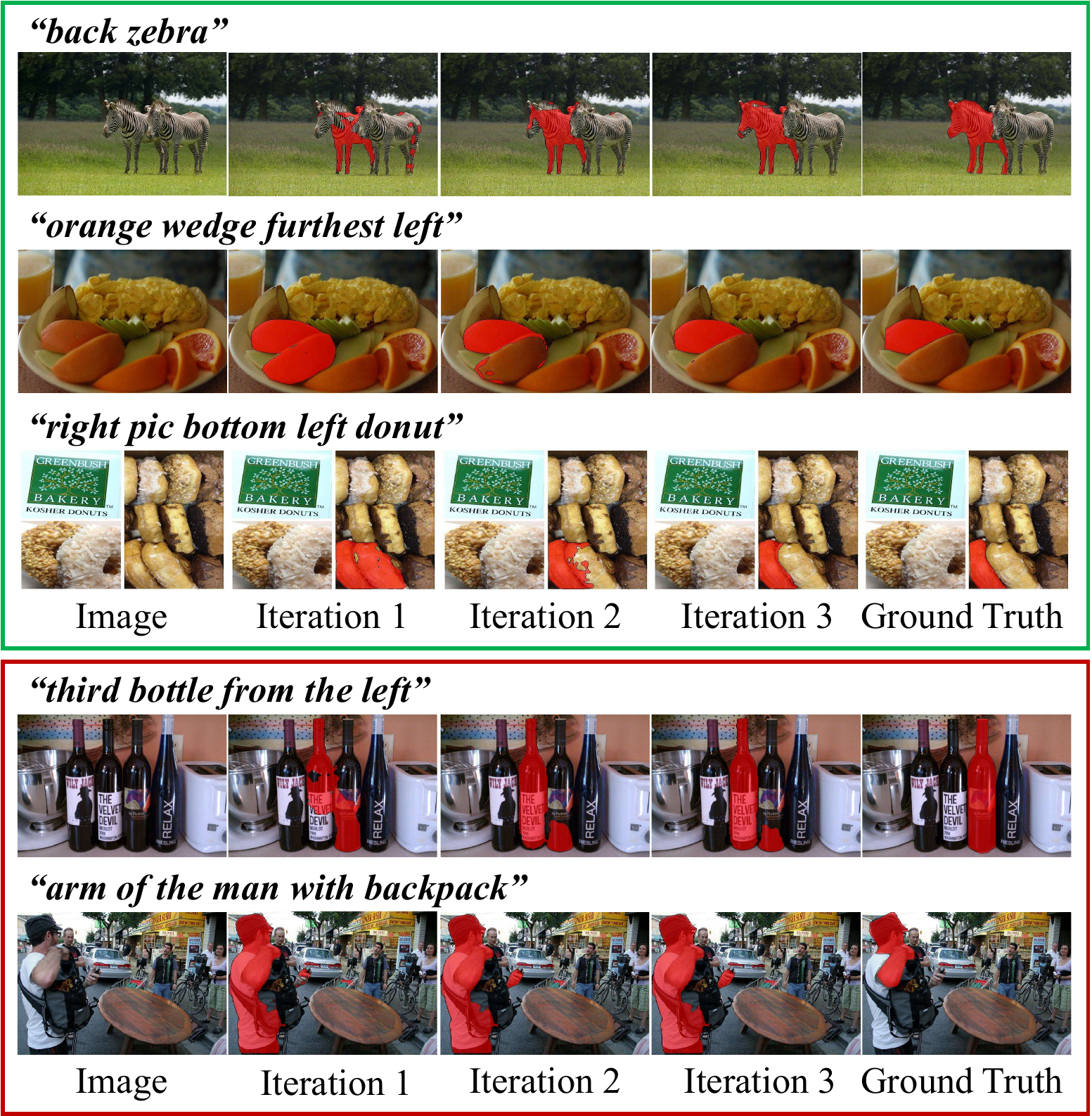}
\caption{Visualizations of predictions from each iteration of SADLR (our default 3-iteration model). Inside the green box are the success cases and inside the red box are the failure cases. In the success cases, the first iteration predicts a mask that roughly locates the intended object and the following iterations correct the missing and/or extra parts progressively.
And the failure cases demonstrate that refinement can fail when the prediction from the first iteration contains too much noise---that is, when localization is too bad. This is a fundamental challenge and solving it requires still more effort devoted to addressing cross-modal understanding.}
\label{fig:5}
\end{figure}

\section{Conclusion}
\label{sec:conclusion}

In this paper, we have proposed a semantics-aware dynamic localization and refinement (SADLR) method for the referring image segmentation problem, which iteratively refines multi-modal feature maps based on aggregated object context.
Extensive experiments on three baseline methods demonstrate the effectiveness and generality of the proposed method.
And evaluations on three standard benchmarks demonstrate its advantage with respect to the state-of-the-art methods.
We hope that our method could inspire further interest in the development of iterative learning approaches for referring image segmentation and be extended to related tasks in the future.

\section*{Acknowledgments}
This work is supported by the UKRI grant: Turing AI Fellowship EP/W002981/1, EPSRC/MURI grant: EP/N019474/1, Shanghai Committee of Science and Technology, China (Grant No. 20DZ1100800), National Natural Science Foundation of China (Grant No. 62206153), HKU Startup Fund, HKU Seed Fund for Basic Research, and Shanghai AI Laboratory.
We would also like to thank the Royal Academy of Engineering and FiveAI.

\bibliography{main.bib}

\appendix
\section*{Appendix}
\label{sec:appendix}
\begin{figure*}[ht]
\centering
\includegraphics[width=\textwidth]{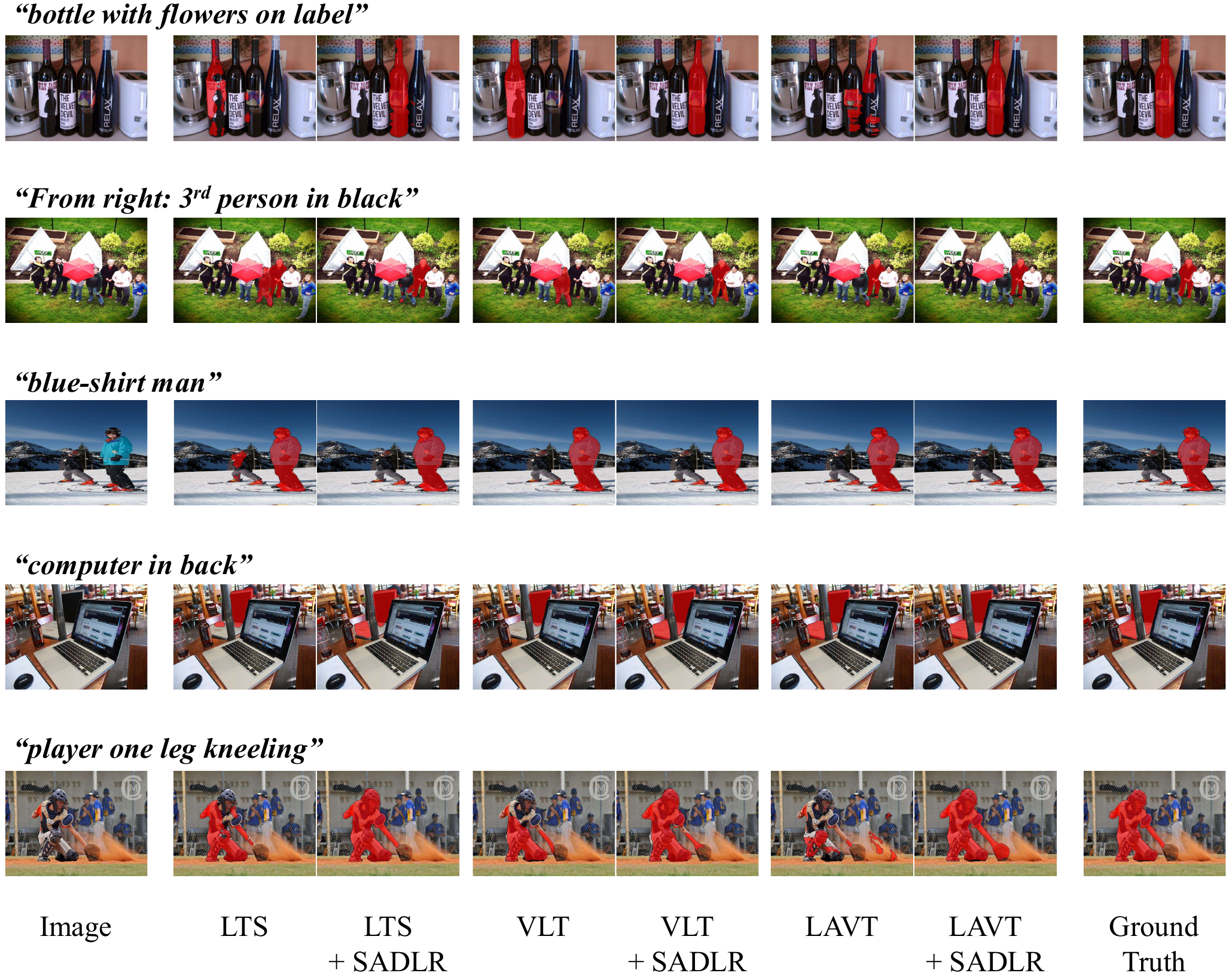}
\caption{Qualitative comparison between a vanilla model (one of LTS, VLT, and LAVT) and its SADLR-enhanced version. For each model, a notable improvement in prediction quality can be observed after the SADLR application. In some examples, wrong localization is able to be corrected (\eg, for LTS and VLT in row 1, and VLT in row 2), and in other examples, segmentation is improved to a high level of accuracy.}
\label{fig:6}
\end{figure*}

\begin{figure*}[ht]
\centering
\includegraphics[width=\textwidth]{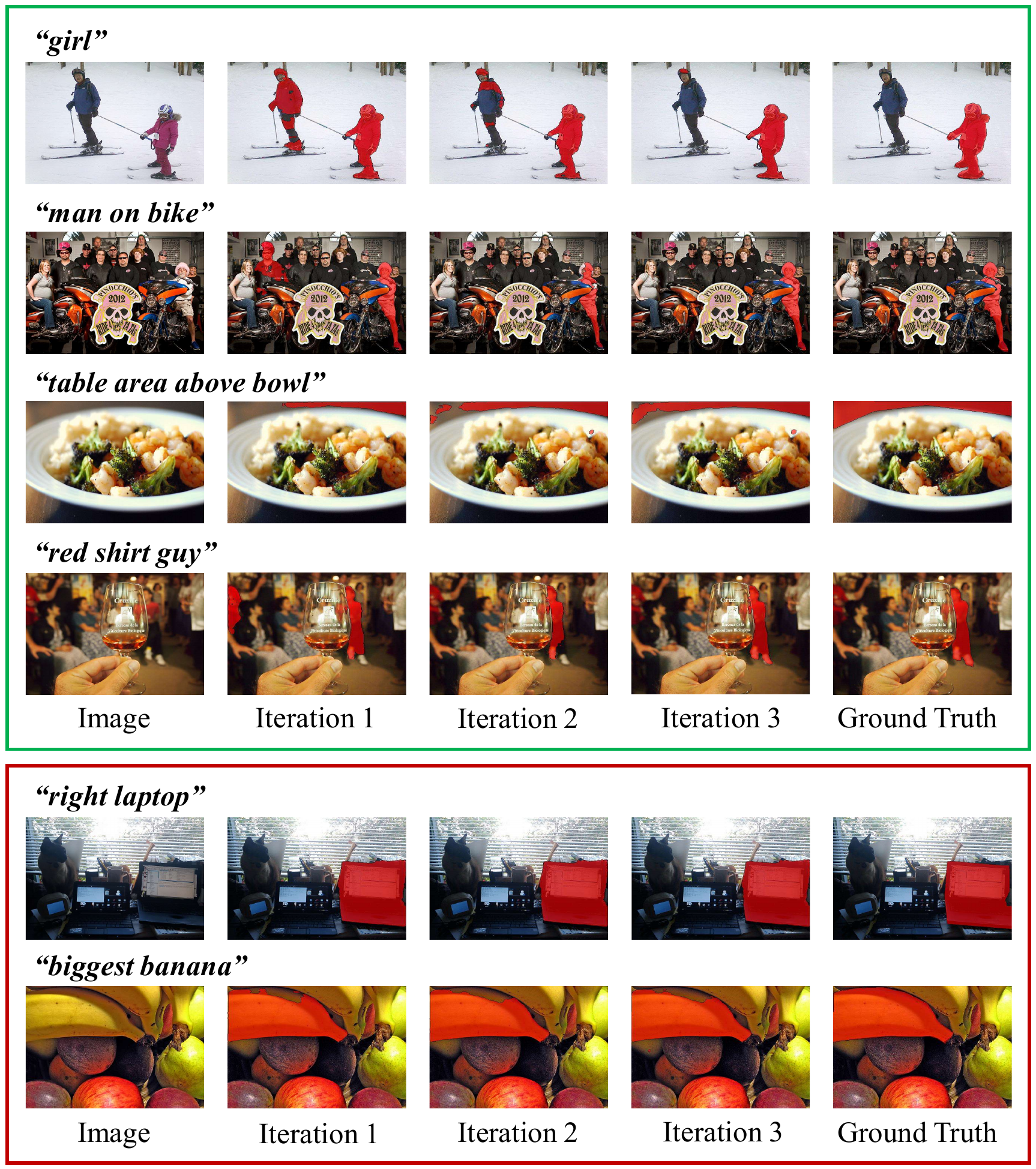}
\caption{Visualizations of predictions from different iterations of SADLR. Examples inside the green box showcase success cases, where the segmentation improves with each iteration. Examples inside the red box illustrate some failure cases. In both examples, some small prediction inaccuracies (false positives or false negatives) near the object boundary are persistent and not refined away. In general, when the object blends in really well with its surroundings, errors along the indistinct parts of its boundary can be hard to correct, and if some factors such as high contrast create spurious signals of separation, then false negatives are, in general, hard to eliminate. We leave the issues above to future work.}
\label{fig:7}
\end{figure*}

We include two supplementary figures in this section.
On the next page, Fig.~\ref{fig:6} visualizes some examples of predictions from the baseline models (\ie, LTS \cite{jing2021locate}, VLT \cite{ding2021vlt}, and LAVT \cite{yang2021lavt}) in comparison with the predictions from our approach.
On the page after the next one, Fig.~\ref{fig:7} illustrates some more success cases and failure cases with predictions from each iteration, supplementing those in Fig.~\ref{fig:5}.

\end{document}